\documentclass{article}

\usepackage[nonatbib,preprint]{neurips_2021_modified}
\usepackage[numbers,sort&compress]{natbib}

\usepackage[utf8]{inputenc} 
\usepackage[T1]{fontenc}    
\usepackage{hyperref}       
\usepackage{url}            
\usepackage{booktabs}       
\usepackage{amsfonts}       
\usepackage{nicefrac}       
\usepackage{microtype}      
\usepackage{wrapfig}
\usepackage{slashbox}
\usepackage{amsmath}

\usepackage[usenames,dvipsnames]{xcolor}
\usepackage{times}
\usepackage[inline]{enumitem}
\usepackage{hyperref}
\usepackage{balance}
\newcommand{\textpapertitle}[1]{{\emph{#1}}}
\balance
\hypersetup{ %
	pdfauthor={Adish Singla},
	pdftitle={},
	pdfsubject={},
    pdflang={en},
	bookmarks,
	bookmarksnumbered,
	backref=page,
	pageanchor=true,
	plainpages=false,
	colorlinks=true,%
	linktocpage=true,
	citecolor=MidnightBlue,%
	filecolor=BrickRed,%
	linkcolor=NavyBlue,%
	urlcolor=NavyBlue 
}

\makeatletter

\newcommand{\Rmnum}[1]{\expandafter\@slowromancap\romannumeral #1@}
\makeatother

\allowdisplaybreaks

\title{
		Reinforcement Learning for Education:\\Opportunities and Challenges\\
		\vspace{4mm}
		\small{\textnormal{Overview of the RL4ED workshop at EDM 2021 conference\thanks{\url{https://rl4ed.org/edm2021/}}}}		
}

\author{
    \ \ \ \ \ \ Adish Singla\\
    \ \ \ \ \ \ MPI-SWS\\
    \ \ \ \ \ \ \texttt{adishs@mpi-sws.org}
	\And
    \ \ \ \ Anna N. Rafferty\\
    \ \ \ \ Carleton College\\
    \ \ \ \ \texttt{arafferty@carleton.edu}	
	\And
    Goran Radanovic\\
    MPI-SWS\\
    \texttt{gradanovic@mpi-sws.org} 
    \And
    Neil T. Heffernan\\
    Worcester Polytechnic Institute\\
    \texttt{nth@wpi.edu}
}

\begin{document}

\maketitle

\begin{abstract}
This survey article has grown out of the RL4ED workshop organized by the authors at the Educational Data Mining (EDM) 2021 conference. We organized this workshop as part of a community-building effort to bring together researchers and practitioners interested in the broad areas of reinforcement learning (RL) and education (ED).  This article aims to provide an overview of the workshop activities and summarize the main research directions in the area of RL for ED.  
\end{abstract}

%


\section{Introduction}
\label{sec:intro}
Reinforcement learning (RL) is a computational framework for modeling and automating goal-directed learning and sequential decision-making~\cite{Puterman1994,sutton2018reinforcement}. Unlike other learning approaches, namely supervised learning and unsupervised learning, RL emphasizes learning by an agent from direct interaction with its environment.  RL is particularly suitable for settings that involve an agent who needs to learn a policy on what to do in different situations---how to map states to actions---to maximize a long-term utility.
The agent must explore different actions to discover actions yielding high reward; crucially, actions affect not only the immediately received reward but also the next state and, through that, all future rewards. These characteristics---actions having long-term consequences, delayed reward, and sequential decision-making under uncertainty---are the key features of RL.

So far, some of the most impressive applications of RL have been limited to game playing only~\cite{mnih2013playing,mnih2015human,silver2016mastering,silver2017mastering}. Given the centrality of sequential student-teacher interactions in education (ED), there has been a surge of interest in applying RL to improve the state-of-the-art technology for ED. There are several problem settings in ED where RL methodology is useful, including training instructional policies using RL methods and modeling a human student using RL.
While promising, it is typically very challenging to apply out-of-the-box RL methods to ED. Further, many problem settings in ED have unique challenges that make the current RL methods inapplicable. Some of the key challenges in ED include the following: (a) the lack of simulation-based-environments to train data-hungry RL methods, (b) the need for large (often unbounded) state space representations, (c) the limited observability of the environment's state (i.e., the student's knowledge), (d) significantly delayed and noisy outcome measures, and (e) concerns about the robustness, interpretability, and fairness of RL methods when applied to the critical domain of ED.



The goal of the RL4ED workshop has been to facilitate tighter connections between researchers and practitioners interested in the broad areas of RL and ED. The workshop focused on two thrusts:
\begin{itemize}[parsep=5pt, leftmargin=*,labelindent=0pt]
\item \textbf{RL\ensuremath{\rightarrow}ED}: Exploring how we can leverage recent advances in RL methods to improve the state-of-the-art technology for ED.
\item \textbf{ED\ensuremath{\rightarrow}RL}: Identifying unique challenges in ED that are beyond the current methodology, but can help nurture technical innovations and next breakthroughs in RL.
\end{itemize}

\section{Overview of the RL4ED@EDM'21 Workshop Activities}
\label{sec:workshop_summary}
\looseness-1In this section, we provide an overview of the RL4ED workshop organized at the EDM 2021 conference; full details are available on the \href{https://rl4ed.org/edm2021/}{workshop website}. The workshop was organized as an online event; we had over $120$ people register and over $60$ simultaneous attendees at the peak. The workshop was structured around invited talks, contributed papers, spotlight presentations, and two panels.

\subsection{Topics of Interests}
As mentioned above, the workshop focussed on two thrusts,  each covering several topics of interest. These topics served as guidelines when selecting the speakers for invited talks and when selecting the contributed papers for spotlight presentations. 

The topics in \textbf{RL\ensuremath{\rightarrow}ED} thrust focussed on leveraging recent advances in RL methods for ED problem settings, including: (i) survey papers summarizing recent advances in RL with applicability to ED; (ii) developing toolkits, datasets, and challenges for applying RL methods to ED; (iii) using RL for online evaluation and A/B testing of different intervention strategies in ED; and (iv) novel applications of RL for ED problem settings.


\looseness-1The topics in \textbf{ED\ensuremath{\rightarrow}RL} thrust focussed on unique challenges in ED problem settings for nurturing next breakthroughs in RL methods, including: (i) using pedagogical theories to narrow the policy space of RL methods; (ii) using RL framework as a computational model of students in open-ended domains; (iii) developing novel offline RL methods that can efficiently leverage historical student data; and (iv) combining statistical power of RL with symbolic reasoning to ensure the robustness for ED.


\subsection{Invited Talks and Panel Sessions}
We invited a set of people from academia and industry to cover various topics of interest and achieve a balance across different perspectives and disciplines.  In total, the workshop had $7$ invited talks; each of these talks being about $25$ minutes long. The final list of speakers, along with their talk titles, is provided below:

\begin{enumerate} [label={[T\arabic*]}, parsep=2pt, leftmargin=*,labelindent=-4pt]
	\item \href{https://people.epfl.ch/tanja.kaeser/?lang=en}{Tanja K{\"a}ser}; \textpapertitle{Modeling and Individualizing Learning in Open-Ended Learning Environments}.
	\item \href{https://www.linkedin.com/in/simon-woodhead/}{Simon Woodhead}; \textpapertitle{Eedi and the NeurIPS 2020 Education Challenge Dataset}.
	\item \href{https://jmhl.org/}{Jos{\'e} Miguel Hern{\'a}ndez Lobato}; \textpapertitle{Deconfounding Reinforcement Learning in Observational Settings}.
	\item \href{https://people.engr.ncsu.edu/mchi/}{Min Chi}; \textpapertitle{The Impact of Pedagogical Policies on Student Learning - A Reinforcement Learning Approach}.
	\item\href{https://cs.stanford.edu/people/ebrun/}{Emma Brunskill}; \textpapertitle{More Practical Reinforcement Learning Inspired by Challenges in Education and Other Societally-Focussed Applications}.
	\item \href{https://psych.wisc.edu/staff/austerweil-joe/}{Joe Austerweil}; \textpapertitle{Is Reinforcement Just a Value to be Maximized?}.
	\item \href{https://faculty.sites.uci.edu/doroudis/}{Shayan Doroudi}; \textpapertitle{Reinforcement Learning for Instructional Sequencing - Learning from Its Past to Meet the Challenges of the Future}.
\end{enumerate}

The video recordings of these invited talks are available on the workshop website.  In addition to these invited talks, the speakers also participated in two separate panel discussions, each of $30$ minutes duration. The Q/A time after the talks and these panel sessions provided an ample opportunity for discussions among workshop participants.

\subsection{Contributed Papers and Spotlight Presentations}
Given the workshop's focus on community-building and networking, we experimented a bit in our call for papers and solicited submissions of two types. The first type, what we called the ``Research track'',  includes papers reporting the results of ongoing or new research, which have not been published before. The second type, what we called the ``Encore track'', includes papers that have been recently published or accepted for publication in a conference or journal. 

For the ``Research track'', we received $4$ submissions and accepted $3$ for the workshop. For the ``Encore track'', we sent invitations to authors with recently published that were relevant to the workshop and we received $6$ submissions in this track. In total, we had $9$ contributed papers covering various topics of interest for the workshop. These contributed papers were presented as spotlight presentations; each of these talks being about $8$ minutes long. In total, we had $10$ spotlight presentations, corresponding to these contributed papers and an additional invited presentation, listed below:

\begin{enumerate} [label={[S\arabic*]}, parsep=2pt, leftmargin=*,labelindent=-4pt]
	\item\textpapertitle{Statistical Consequences of Dueling Bandits}. \\ (Research track; \cite{RL4ED_EDM2021_Saxena})
	\item \textpapertitle{Capturing Student-Robot Interactions for a Data-Driven Educational Dialogue RL Environment}.  \\ (Research track; \cite{RL4ED_EDM2021_Maidment})
	\item \textpapertitle{Towards Transferrable Personalized Student Models in Educational Games}.  \\ (Encore track; \cite{DBLP:conf/atal/SpauldingSPB21})
	\item \textpapertitle{Extending Adaptive Spacing Heuristics to Multi-Skill Items}. \\ (Encore track; \cite{RL4ED_EDM2021_Choffin})
	\item \textpapertitle{Getting Too Personal(ized): The Importance of Feature Choice in Online Adaptive Algorithms}. \\ (Encore track; \cite{DBLP:conf/edm/LiYSSWR20})
	\item  \textpapertitle{Approximately Optimal Teaching of Approximately Optimal Learners}.  \\ (Encore track; \cite{DBLP:journals/tlt/WhitehillM18})
	\item \textpapertitle{Learning Expert Models for Educationally Relevant Tasks using Reinforcement Learning}. \\ (Encore track; \cite{Maclellan_EDM2021})
	\item \textpapertitle{Deep Reinforcement Learning to Simulate, Train, and Evaluate Instructional Sequencing Policies}.  \\ (Research track; \cite{RL4ED_EDM2021_Subramanian})
	\item \textpapertitle{Adaptively Scaffolding Cognitive Engagement with Batch Constrained Deep Q-Networks}. \\ (Encore track; \cite{DBLP:conf/aied/FahidRSGPL21})
	\item \textpapertitle{Integrating Reinforcement Learning into the ASSISTments Platform}.  \\ (Additional invited spotlight presentation)
\end{enumerate}

The video recordings of these spotlight presentations are available on the workshop website. 

\section{Summary of the Main Research Directions in RL4ED}
\label{sec:research_directions}

In this section, we summarize the main research directions in the area of RL for ED.

\paragraph{RL methods for personalizing curriculum across tasks.} The most direct and well-studied application of RL for ED is to train an instructional policy for providing students with a personalized curriculum. In this problem setting, one trains an RL agent to induce an instructional policy in an intelligent tutoring system, and the human student is part of the ``environment'' as per RL terminology~\cite{sutton2018reinforcement}. For a given student, such an instructional policy maps the student's history of responses to the next task to maximize long-term learning gains. We refer the reader to \cite{doroudi2019s}, which provides an excellent survey on this topic; also, see invited talk [T7] from the authors of the survey. The latest research in this direction is also covered by the invited talk [T4], spotlights [S6] and [S8], as well as several recent works~\cite{DBLP:journals/cogsci/RaffertyBGS16,DBLP:conf/aied/SawyerRL17,DBLP:journals/tlt/WhitehillM18,DBLP:conf/edm/AusinABC19,zhou2019hierarchical,DBLP:conf/chi/BassenBSTPZGFM20}.
Despite being a natural application for RL, in practice, there are several challenges to train effective RL-based policies for ED domains in the real world; see \cite{doroudi2019s}. One of the major challenges is that the students' true knowledge state is not directly observable~\cite{DBLP:journals/cogsci/RaffertyBGS16,DBLP:journals/tlt/WhitehillM18} -- we need to use appropriate representations for mapping students' responses to a knowledge state. Another major challenge arises from the lack of simulation-based environments to train data-hungry RL methods. An RL agent typically requires millions of training episodes, and such training is generally done in a simulator for gaming domains; however, we do not have such realistic simulators or computational models of human students for ED domains.  To tackle these challenges, an important research direction is to investigate how to effectively combine policies derived from RL methods and pedagogical theories, or using pedagogical theories to narrow the policy space of RL for ED problem settings. Another important research direction is developing novel offline RL methods that can efficiently leverage historical data; see invited talk [T5].

\paragraph{RL methods for providing hints, scaffolding, and quizzing.} Beyond curriculum across tasks, another important application of RL for ED is to train policies that can provide hints as feedback within a task. Especially for complex open-ended domains, such as block-based visual programming or high-school algebra, hints feedback and scaffolding play an important role in improving student engagement and learning gains~\cite{aleven2016instruction,DBLP:conf/lats/WilliamsKRMGLH16,DBLP:conf/lats/PatikornH20,DBLP:conf/lak/EricksonBMVH20}. In one of the early works, \cite{barnes2008toward} used Markov Decision Processes formalism for automatic generation of hints for logic proof tutoring using historical student data. In recent contemporary work, \cite{edm20-zero-shot} proposed an RL framework to train a hints policy for block-based visual programming tasks without relying on historical student data, thereby tackling the zero-shot challenge of providing hints in this domain. We also refer the reader to recent works of \cite{DBLP:conf/edm/JuCZ20,DBLP:conf/aied/FahidRSGPL21}
 where RL methods are used for scaffolding and assisting at critical decision-making points; also see invited talk [T4] and spotlight [S9] from the authors of these papers. In a somewhat different problem setting, \cite{edm21-quizzing-policy} have investigated how an RL-based policy can be used for quizzing students to infer their knowledge state, tackling the challenge\#4 in Eedi's NeurIPS Education Challenge (see \cite{eedi,wang2020diagnostic} and invited talk [T2]).  
These works are still in the early stages and showcase the potential of using RL-based policies for different problem settings beyond curriculum across tasks. In the coming years, we believe that RL methods can play a critical role in providing feedback to students for complex, open-ended tasks. An exciting direction of research here would be to train RL-based policies that can balance different objectives when providing hints -- completing the current task quickly with hints vs. maximizing the pedagogical value of hints in terms of the learner's knowledge gain (i.e., better performance in future tasks).

\paragraph{RL for adaptive experimentation and A/B testing in educational platforms.} In recent years, there is a growing interest in using RL methods for assessing different educational interventions in large-scale online platforms. In particular, a special family of RL methods, called multi-armed bandits (MAB), are used for adaptive experimentation in recent works -- each student is assigned to a version of the technology or a type of intervention (aka ``arm'' in MAB terminology), and the algorithm observes the student's learning outcome (the reward associated with the assigned ``arm''); each subsequent student is more likely to be assigned to a version of the technology that has been more effective for previous students, as the algorithm discovers what is effective~\cite{DBLP:conf/edm/LiuMBP14,DBLP:conf/chi/WilliamsRTALK18,DBLP:conf/aied/RaffertyYW18}. While the standard MAB algorithms do not allow per-student personalization, the contextual MAB algorithms can also account for student features and personalize the assignment to further improve learning gains. In recent work, \cite{DBLP:journals/tlt/WhitehillM18} investigated the effect of features used in contextual MAB algorithms and highlighted the trade-offs of personalization on learning gains (also, see spotlight [S5] from the authors of this paper). In another recent work, \cite{DBLP:conf/aied/Zavaleta-Bernuy21} conducted adaptive experiments as a case study of sending homework email reminders to students, and the paper reports various open issues that arise in conducting such experiments in real-world settings. We also refer the reader to several other recent works, including \cite{rafferty2019statistical,DBLP:conf/aied/MuiLD21} as well as spotlights [S1] and [S10]. In [S10], the authors discuss their ongoing effort to bring MAB-based adaptive experimentation to recommend and personalize the content students receive in the ASSISTments educational platform~\cite{ASSISTments}. Overall, RL for adaptive experimentation is a very promising area where we expect to see the deployment of RL-driven techniques in real-world educational platforms in the coming years. An important research direction in this area is to better understand the ethical implications of adaptive experiments, and design contextual MAB algorithms that can account for fairness and ensure educational equity across different groups.



\paragraph{RL framework to model a human student.} In contrast to using the RL agent for representing a teacher / tutoring system, one can take an orthogonal view and use the RL framework for modeling the students' learning or problem-solving process.  In this setting, the human student is modeled as an RL agent, and the teacher represents the ``environment''; cf. the setting of training an instructional policy where the RL agent represents the teacher or tutoring system.  This modeling framework is especially useful in open-ended learning domains where tasks are conceptual, open-ended, and sequential, including domains such as block-based visual programming and high-school algebra.  Such an RL-based computational model of human students is useful for a variety of settings. For instance, one could use such a model to diagnose students' mistakes based on their attempted solutions and in designing more effective environment feedback (e.g., via appropriate interventions); see \cite{rafferty2020assessing,DBLP:conf/edm/YangZTAC20}. Furthermore, such a computational model could be used as simulated students to evaluate teaching algorithms or to train instructional policies. In the machine teaching research \cite{DBLP:journals/corr/abs-1801-05927}, a series of recent works have used RL-based agents as a student model to investigate the theoretical foundations of teaching for sequential decision-making tasks. For instance, \cite{DBLP:conf/nips/HaugTS18,DBLP:conf/nips/TschiatschekGHD19,DBLP:conf/ijcai/KamalarubanDCS19} studied the problem of curriculum design and optimizing demonstrations when the student is modeled as an imitation learning agent, and \cite{DBLP:conf/icml/RakhshaRD0S20}  studied the problem of policy teaching and environment design when the student is modeled as an RL agent.  A number of recent works and workshop activities focussed on this research direction, including \cite{mcilroy2020aligning,DBLP:conf/atal/SpauldingSPB21,Maclellan_EDM2021}, invited talks [T1] and [T6], and spotlights [S3] and [S7]. In the coming years, we believe that the RL framework to model a human student will continue to be an important research direction. One of the most important research questions is how to incorporate human-centered aspects of learning into an RL agent so that these agents are a better representation of human students. More concretely, it would be important to develop RL agents that can capture the capabilities of human learners, e.g., few-shot learning, deductive reasoning, and learning from different feedback types.

\paragraph{RL for educational content generation.} Another important research direction is to leverage RL methodology for educational content generation, such as generating new exercises, quizzes, or videos.  Often referred to as procedural content generation (PCG), recent works have explored the applicability of RL for PCG in the context of generating different levels of Sokoban puzzles~\cite{khalifa2020pcgrl,kartal2016data} and racing games~\cite{gisslen2021adversarial}. In recent work, \cite{ahmed20synthesizing} used Monte Carlo Tree Search (MCTS) method combined with symbolic techniques to synthesize new tasks for block-based visual programming domain. These synthesized tasks could be useful in many different ways
in practical systems -- for instance, tutors can assign new practice tasks as homework or quizzes to students to check their knowledge, and students can automatically obtain new similar tasks after
they failed to solve a given task. Given the critical need to provide personalized and diverse educational content on online platforms, RL for educational content generation is an important research area that needs to be explored further.


\section{Conclusions}
\label{sec:conclusions}

RL for ED is an important application area for future work that may lead to practical improvements in education and to new advances in reinforcement learning. The talks and discussions at the EDM2021 workshop highlighted excitement in the community around the main areas covered in this document, and the diverse array of perspectives as well as panelist comments demonstrated the importance of drawing on ideas from multiple disciplines, including (but not limited to) the learning sciences, cognitive science, and machine learning. This need for multiple perspectives and the unique challenges raised by educational applications suggest the need for continued fostering of community in this area.

\bibliography{references}
\bibliographystyle{unsrt}

\end{document}